\def\year{2019}\relax
\documentclass[letterpaper]{article} 
\usepackage{aaai19}  
\usepackage{times}  
\usepackage{helvet}  
\usepackage{courier}  
\usepackage{url}  
\usepackage{graphicx}  
\usepackage{amsmath}
\usepackage{amsfonts}
\usepackage{subfig}
\usepackage{algorithm}
\usepackage{algpseudocode}

\begin{document}
%
\title{Geometric Multi-Model Fitting by Deep Reinforcement Learning}
\author{Zongliang Zhang, Hongbin Zeng, Jonathan Li*, Yiping Chen, Chenhui Yang, Cheng Wang\\
Fujian Key Laboratory of Sensing and Computing for Smart Cities, School of Information Science and Engineering\\
Xiamen University, Xiamen, Fujian 361005, China (*Corresponding author)\\
\{zhangzongliang, hbzeng\}@stu.xmu.edu.cn, \{junli, chenyiping, chyang, cwang\}@xmu.edu.cn\\
}
\maketitle
\begin{abstract}
This paper deals with the geometric multi-model fitting from noisy, unstructured point set data (e.g., laser scanned point clouds). We formulate multi-model fitting problem as a sequential decision making process. We then use a deep reinforcement learning algorithm to learn the optimal decisions towards the best fitting result. In this paper, we have compared our method against the state-of-the-art on simulated data. The results demonstrated that our approach significantly reduced the number of fitting iterations.
\end{abstract}


\section{Introduction}
Geometric model fitting aims to reconstruct underlying models (e.g., lines, circles, characters, and buildings) from given data (e.g., images or laser scanning point clouds). With the reconstructed rich model information (e.g., shape, scale, rotation, and location), the data can be comprehensively understood. With such merit, model fitting has constantly attracted research interests for a long time. However, the model fitting problem is far from being solved, at least in terms of computational speed, because of increasing complexity of encountered data and thus models. A common case of complex data is that data conceive multiple models. For example, a CAPTCHA image usually contains multiple characters \cite{george2017generative}. A multi-model fitting technique is needed to handle such data.

A recent trend for addressing model fitting problem is to formulate it as an optimization problem \cite{lake2015human}, such that it can be conveniently tackled by an existing optimization algorithm. Our previous method \cite{zhang2019robust} uses the cuckoo search (CS) algorithm \cite{yang2010engineering} to solve the optimization problem in model fitting, and notably achieves the new state-of-the-art in the challenging few-shot character recognition tasks (George et al. 2017).	CS can approach the optimum with high precision. However, it usually takes many iterations to converge to the optimum, especially when the fitting involves a large number of variables, which is the case of multi-model fitting. The number of variables involved in $n$-model fitting is as large as $n$ times that in single-model fitting. In other words, it is time-consuming to use CS to perform multi-model fitting. 

In this paper, we propose a reinforcement learning approach for optimization in multi-model fitting. Our insight is as follows. The selection of variable values for a model can be seen as a decision. The fitting of multiple models is a process that consists of a sequence of decisions. Such decision making process can be efficiently optimized by reinforcement learning.

The work similar to ours can be found in \cite{teboul2013parsing}, which uses a traditional reinforcement learning method to control binary split shape grammar for parsing facade images. Their method works under the assumption that the split grammar has discrete variables. However, the variables involved in model fitting usually are continuous, which are challenging for a traditional reinforcement learning method to handle \cite{lillicrap2015continuous}. In contrast, our work is based on recently developed deep reinforcement learning (DRL), which has made remarkable progress for a number of challenging tasks including continuous control \cite{lillicrap2015continuous}. 

\section{Method}

A geometric model $M$ is a $k$-dimensional point set, i.e., $ M \subset \mathbb{R}^k$. In this paper, $k=3$. Give a data point set $D \subset \mathbb{R}^k$, the goal of model fitting is to find a model $M$ that is most similar to $D$. For multi-model fitting, $M$ is the union set of multiple models, i.e., $M = \bigcup\nolimits_{i = 1}^n {M_g^{{\theta _i}}} $, where $n$ is the number of models, and ${M_g^{{\theta }}}$ is the model defined by a given parametric rule $g$ which is parameterized by variable $\theta$. Formally, multi-model fitting can be formulated as the following maximization problem:
\begin{equation}\label{eq:maximization}
\mathop {\max }\limits_{({\theta _1},{\theta _2}, \cdots ,{\theta _n})} f({\theta _1},{\theta _2}, \cdots ,{\theta _n}) = s(\bigcup\nolimits_{i = 1}^n {M_g^{{\theta _i}}} ,D),
\end{equation}
where $s(\cdot,\cdot)$ is the geometric similarity estimator defined in \cite{zhang2019robust}. The pseudo-code of our method is shown in Algorithm \ref{alg:method}, where the DRL actor $q$ is based on \cite{lillicrap2015continuous}, and $\mathcal{N}_i$ is exploration noise \cite{lillicrap2015continuous}. Our method follows a hypothesis and verify paradigm to solve the maximization problem Eq. \eqref{eq:maximization}. In each iteration, for each model, a hypothesis value $\theta _i$ is proposed according to the actor and exploration noise. Then the hypothesis is verified through computing the reward in order to update the actor.

\begin{algorithm}[tb!]
\caption{The proposed method}
\label{alg:method}
\begin{algorithmic}
\State\textbf{input}: a data point set $D$, a parametric rule $g$ with variable ${\theta}$, the number of models $n$, the DRL actor $q$, the verify function $f$
\State\textbf{output}: $({\theta _1^*},{\theta _2^*}, \cdots ,{\theta _n^*})$ that maximizes Eq. \eqref{eq:maximization}
\State Randomly initialize $({\theta _1^*},{\theta _2^*}, \cdots ,{\theta _n^*})$
\For {iteration $j=1$ to $j_{max}$}
\For {$i = 1$ to $n$}
\State Select variable $\theta _i = q(i) + \mathcal{N}_i$
\State Observe reward: 
\State \;\;\;\;\;\; $r_i=f({\theta _1},{\theta _2}, \cdots ,{\theta _i})-f({\theta _1},{\theta _2}, \cdots ,{\theta _{i-1} })$
\State Update $q$ according to $r_i$
\EndFor
\If {$ f(q(1),q(2),\cdots,q(n))>f({\theta _1^*},{\theta _2^*}, \cdots ,{\theta _n^*})$} \[(\theta _1^*,\theta _2^*, \cdots ,\theta _n^*) = \left( {q(1),q(2), \cdots ,q(n)} \right)\]
\EndIf
\EndFor
\end{algorithmic}
\end{algorithm}

We now present the computation cost of the proposed DRL based $n$-model fitting method. In each iteration, the computational time cost of a hypothesis and verify algorithm is composed of two parts: the hypothesis part $t_\text{DRL}^\text{H}$ and the verify part $t_\text{DRL}^\text{V}$. Therefore, the total computational cost is $t_\text{DRL}=j_\text{DRL} (t_\text{DRL}^\text{H}+ t_\text{DRL}^\text{V})$, where $j_\text{DRL}$ is the total number of iterations. For $n$-model fitting, it is needed to calculate the verify function $f$ for $n$ times in each iteration. Let $t_f$ be the cost to calculate $f$ one time, then $t_\text{DRL}^\text{V}=n t_f$, and $t_\text{DRL}=j_\text{DRL} (t_\text{DRL}^\text{H} + n t_f  )$. In contrast, the CS based method \cite{zhang2019robust} only needs to calculate $f$ one time in one iteration. Consequently, the total computational cost of the CS based method is $t_\text{CS}=j_\text{CS} (t_\text{CS}^\text{H}+ t_f  )$. It can be concluded that, when $t_f  \gg  t_\text{DRL}^\text{H}$ and $ j_\text{CS}>n j_\text{DRL}$, DRL is more efficient than CS. Note that $t_f$ is determined by the data and the model sizes \cite{zhang2019robust}. Therefore, $t_f  \gg  t_\text{DRL}^\text{H}$ holds for many applications in which data and model sizes are large. For example, a laser scanning point cloud is usually large in size as containing millions of points.

\section{Experiments}
\begin{figure}[tb!]
\centering
\subfloat[]{\label{fig:clean-data}\includegraphics[height=0.7in]{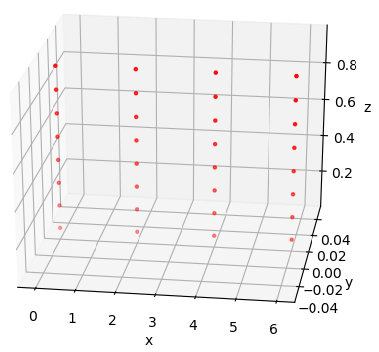}}
\hfill
\subfloat[]{\label{fig:noisy-data}\includegraphics[height=0.7in]{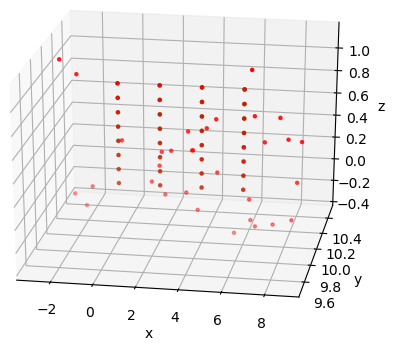}}
\hfill
\subfloat[]{\label{fig:cs}\includegraphics[height=0.7in]{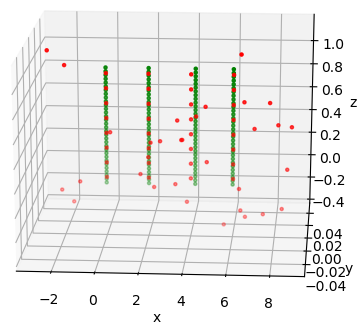}}
\hfill
\subfloat[]{\label{fig:drl}\includegraphics[height=0.7in]{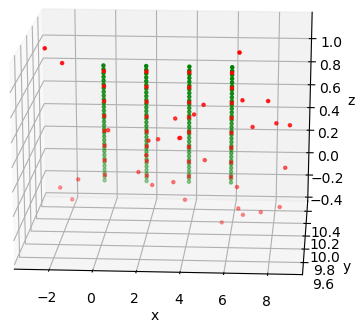}}
\caption{Data point sets and fitted models. \protect\subref{fig:clean-data} Clean data $D_1$. \protect\subref{fig:noisy-data} Corrupted data $D_2$, which is generated by adding some outliers to $D_1$. \protect\subref{fig:cs} The model fitted by CS after 1000 iterations. \protect\subref{fig:drl} The model fitted by DRL after 100 iterations.}
	\label{fig:data-and-models}
\end{figure}

In this abstract, we preliminarily evaluate our method by fitting line segments to the data $D_2$ shown in Fig. \ref{fig:noisy-data}. Specifically, the parametric rule $g$ input to our method is a vertical line segment rule with only one variable $\theta \in \mathbb{R}$ that determines the horizontal location of the line segment. We also fix the number of models $n=4$.

\begin{figure}[tb!]
\centering
\includegraphics[width=1\columnwidth]{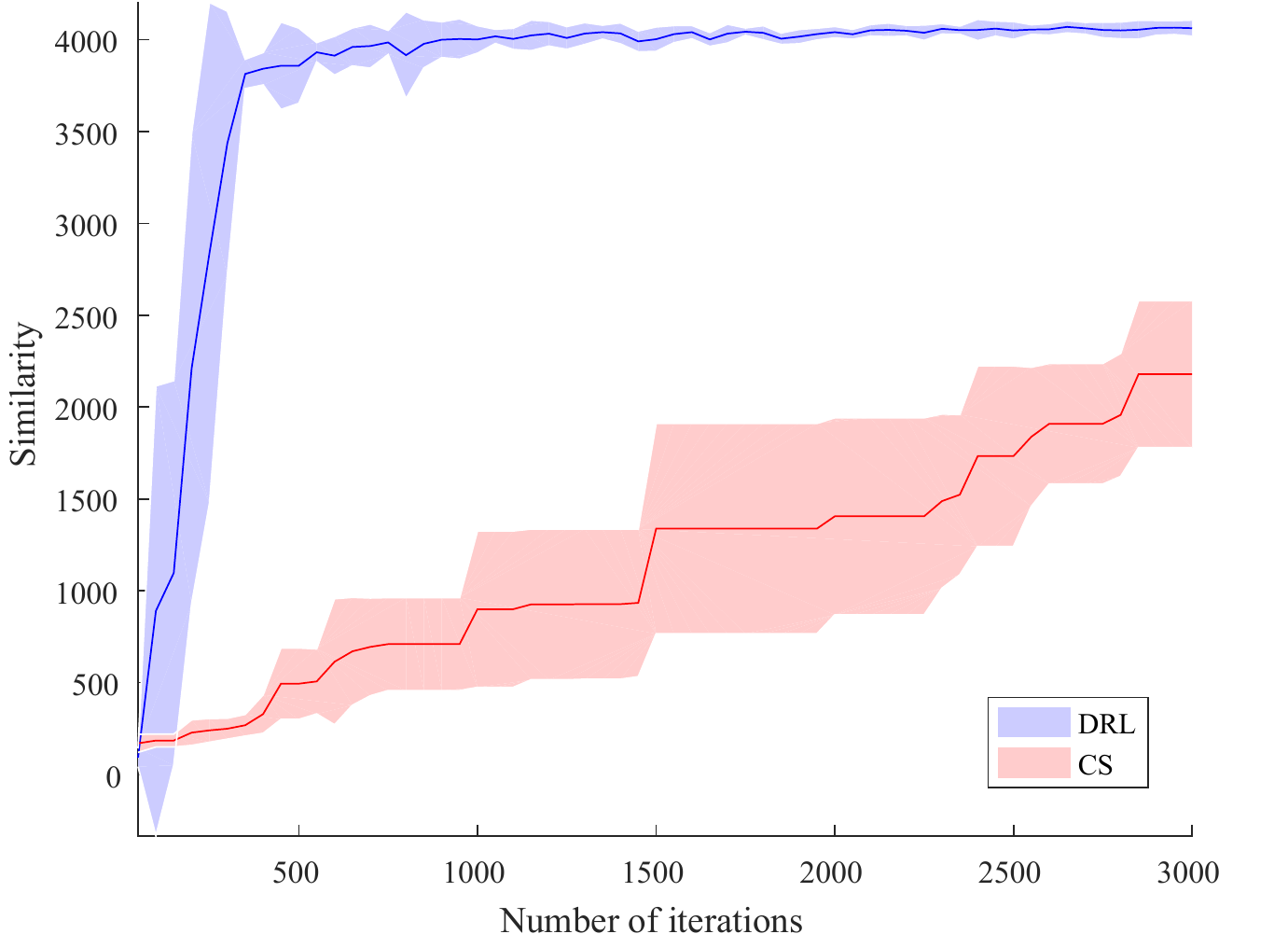}
\caption{Fitting similarities at different iterations of DRL and CS.}
\label{fig:evolution}
\end{figure}

As shown in Figs. \ref{fig:cs} and \ref{fig:drl}, DRL fits the data well after only 100 iterations, whereas CS cannot well fit the data even after 1000 iterations. The evolutions of similarity during fitting are shown in Fig. \ref{fig:evolution}, where each line represents the mean of similarity values and the patches around each line represents the standard deviations. The mean values and standard deviations are computed from 5 times of fitting. The results clearly indicates that DRL is tens of times more efficient than CS in terms of the numbers of fitting iterations.

\bibliographystyle{aaai}
\bibliography{Zhang2019GeometricMMFDRL}
\end{document}